\documentclass[runningheads,a4paper]{llncs}

\usepackage{subfigure}
\usepackage{subfloat}
\usepackage{url}
\usepackage{algorithm}
\usepackage{algorithmicx}
\usepackage{algpseudocode}
\usepackage{graphicx}
\usepackage{paralist}
\usepackage{mdwmath}
\usepackage{mdwtab}
\usepackage{amsmath}
\usepackage{amssymb}
\usepackage{lipsum}
\usepackage{moreverb}
\usepackage{caption}
\usepackage{tabto}
\usepackage[colorlinks,citecolor=black,linkcolor=black, urlcolor=blue]{hyperref}

\newcommand{\argmax}{\operatornamewithlimits{argmax}}

\urldef{\mailsa}\path|{saminda,aseek,visser}@cs.miami.edu|

\graphicspath{{./}{./figures/}{./plots/}{./presentation/figures/}}

\begin{document}

\markboth{Saminda Abeyruwan, Andreas Seekircher, and Ubbo Visser}{Connection Science}

\title{Off-Policy General Value Functions to Represent Dynamic Role Assignments in RoboCup 3D Soccer Simulation}
\titlerunning{~}
\author{Saminda Abeyruwan  \and Andreas Seekircher  \and Ubbo Visser}
%
%\authorrunning{Abeyruwan et al.}
% (feature abused for this document to repeat the title also on left hand pages)

% the affiliations are given next
\institute{Department of Computer Science\\
University of Miami\\
1365 Memorial Drive, Coral Gables, FL, 33146 USA\\
\mailsa
}

\maketitle

\begin{abstract}
Collecting and maintaining accurate world knowledge in a dynamic, complex, adversarial, and stochastic environment
such as the RoboCup 3D Soccer Simulation is a challenging task. Knowledge should be
learned in real-time with time constraints. We use recently introduced Off-Policy Gradient Descent
algorithms within Reinforcement Learning that illustrate learnable knowledge representations for
dynamic role assignments. The results show that the agents have learned competitive policies against the top
teams from the RoboCup 2012 competitions for three vs three, five vs five, and seven vs seven agents. We have explicitly used subsets of agents to identify the dynamics and the semantics for which the agents learn to maximize their performance measures, and to gather knowledge about different objectives, so that all agents participate effectively and efficiently within the group.

\bigskip

\end{abstract}

\section{Introduction}
\label{sec:introduction}
% no \IEEEPARstart
% You must have at least 2 lines in the paragraph with the drop letter
% (should never be an issue)
The \textit{RoboCup 3D Soccer Simulation} environment provides a dynamic, real-time, complex, adversarial, and stochastic
multi-agent environment for simulated agents. The simulated agents formalize their
goals in two layers: \begin{inparaenum}\item the physical layers, where controls related to walking,
kicking etc. are conducted; and \item the decision layers, where high level actions are taken to emerge behaviors. \end{inparaenum}
In this paper, we investigate a mechanism suitable for decision layers to use recently introduced Off-Policy Gradient
Decent Algorithms in Reinforcement Leaning (RL) that illustrate learnable knowledge representations
to learn about \textit{a dynamic role assignment function}.

In order to learn about an effective dynamic role assignment function, the agents need to
consider the dynamics of agent-environment interactions. We consider these interactions as the
agent's knowledge. If this knowledge is represented in a formalized form
(e.g., first-order predicate logic) an agent could infer many aspects about its interactions
consistent with that knowledge. The knowledge representational forms show different degrees of
computational complexities and expressiveness \cite{DBLP:conf/atal/SuttonMDDPWP11}. The
computational requirements increase with the extension of expressiveness of the representational
forms. Therefore, we need to identify and commit to a representational form, which is scalable for
on-line learning while preserving expressivity. A \textit{human} soccer player knows a lot
of information about the game before (s)he enters onto the field and this prior knowledge influences the
outcome of the game to a great extent. In addition, human soccer players dynamically change their
knowledge during games in order to achieve maximum rewards. Therefore, the knowledge of the human
soccer player is to a certain extent either \textit{predictive} or \textit{goal-oriented}. Can a
\textit{robotic} soccer player collect and maintain predictive and goal-oriented knowledge? This is a challenging problem for agents with time constraints
and limited computational resources.

We learn the role assignment function using a framework that is developed based on the concepts of
Horde, the real-time learning methodology, to express knowledge using General Value
Functions (GVFs) \cite{DBLP:conf/atal/SuttonMDDPWP11}. Similar to Horde's sub-agents, the agents in
a team are treated as independent RL sub-agents, but the agents take actions based on their belief of the
world model. The agents may have different world models due to noisy perceptions and communication
delays. The GVFs are constituted within the RL framework. They are predictions or off-policy controls
that are answers to questions. For example, in order to make a prediction a question must be asked
of the form ``If I move in this formation, would I be in a position to score a goal?'', or ``What set
of actions do I need to block the progress of the opponent agent with the number 3?''. The question
defines what to learn. Thus, the problem of prediction or control can be addressed by learning value functions. An
agent obtains its knowledge from information communicated back and forth between the agents and
the agent-environment interaction experiences.

There are primarily two algorithms to learn about the GVFs, and these algorithms are based on Off-Policy Gradient Temporal Difference (OP-GTD) learning: \begin{inparaenum}
                                                                        \item with action-value methods,
a prediction question uses GQ($\lambda$) algorithm \cite{Maei_Sutton_2010}, and a control or a goal-oriented question
uses Greedy-GQ($\lambda$) algorithm \cite{DBLP:conf/icml/MaeiSBS10}. These algorithms learned about a deterministic target policies and the control algorithm finds the greedy action with respect to the action-value function; and
\item with policy-gradient methods, a goal-oriented question can be answered using Off-Policy Actor-Critic algorithm \cite{DBLP:journals/corr/abs-1205-4839}, with an extended state-value function, GTD($\lambda$) \cite{MaeiHRPhdThesis2011}, for GVFs. The policy gradient methods are favorable for problems having stochastic optimal policies, adversarial environments, and problems with large action spaces. \end{inparaenum}
The OP-GTD algorithms possess a number of
properties that are desirable for on-line learning within the RoboCup 3D Soccer Simulation environment: \begin{inparaenum}
\item off-policy updates;
\item linear function approximation;
\item no restrictions on the features used;
\item temporal-difference learning;
\item on-line and incremental;
\item linear in memory and per-time-step computation costs; and
\item convergent to a local optimum or equilibrium point \cite{DBLP:conf/nips/SuttonSM08,DBLP:conf/icml/MaeiSBS10}.
\end{inparaenum}

In this paper, we present a methodology and an implementation to learn about a dynamic role assignment
function considering the dynamics of agent-environment interactions based on
GVFs. The agents ask questions and approximate value functions answer to those questions. The
agents independently learn about the role assignment functions in the presence of an adversary team.
Based on the interactions, the agents may have to change their roles in order to continue in the
formation and to maximize rewards. There is a finite number of roles that an agent can commit to, and
the GVFs learn about the role assignment function. We have conducted all our experiments in the RoboCup 3D
Soccer Simulation League Environment. It is based on the general purpose multi-agent simulator
SimSpark\footnote{\url{http://svn.code.sf.net/p/simspark/svn/trunk/}}. The robot agents in the simulation are modeled
based on the Aldebaran NAO\footnote{\url{http://www.aldebaran-robotics.com/}} robots. Each robot has
22 degrees of freedom. The agents communicate with the server through message passing and each agent
is equipped with noise free joint perceptors and effectors. In addition to this, each agent has a
noisy restricted vision cone of $120^o$. Every simulation cycle is limited to $20~ms$, where
agents perceive noise free angular measurements of each joint and the agents stimulate the necessary
joints by sending torque values to the simulation server. The vision information from the server is
available every third cycle ($60~ms$), which provides  spherical coordinates of the perceived
objects. The agents also have the option of communicating with each other every other simulation
cycle ($40~ms$) by broadcasting a $20~bytes$ message. The simulation league competitions
are currently conducted with 11 robots on each side (22 total).

The remainder of the paper is organized as follows: In Section \ref{sec:RelatedWork}, we briefly
discuss knowledge representation forms and existing role assignment formalisms. In
Section \ref{sec:LearnableknowledgeRepresentationForRoboticSoccer}, we introduce GVFs within the
context of robotic soccer. In
Section \ref{sec:DynamicRoleAssignment}, we formalize our mechanisms of dynamic role assignment
functions within GVFs. In Section \ref{sec:GVFQandA}, we identify the question and answer functions to represent GVFs, and Section \ref{sec:Experiments} presents the experiment results and the discussion. Finally, Section
\ref{sec:ConclusionAndFutureWork} contains
 concluding remarks, and future work.

%%%%%%%%%%%%%%%%%%%%%%%%%%%%%%%%%%%%%%%%%%%%%%%%%%%%%%%%%%%%%%%%%%%%%%%%%%%%%%%%%%%%%%%%%%%%%%%%%%
\section{Related Work}
\label{sec:RelatedWork}

One goal of multi-agent systems research is the investigation of the prospects of efficient cooperation among a set of agents in
real-time environments. In our research, we focus on the cooperation of a set of agents in a
real-time robotic soccer simulation environment, where the agents learn about an optimal or a
near-optimal role assignment function within a given formation using GVFs. This subtask is
particularly challenging
compared to other simulation leagues considering the limitations of the environment, i.e. the limited locomotion capabilities, limited communication bandwidth, or crowd management rules. The role assignment is a part of the hierarchical machine
learning paradigm \cite{AIJ99,Stone99layeredlearning}, where a formation defines the role space.
Homogeneous agents can change roles flexibly within a formation to maximize a given reward function.

RL framework offerers a set of tools to design sophisticated and hard-to-engineer
behaviors in many different robotic domains (e.g., \cite{Bagnell_2013_7451}). Within the domain of \textit{robotic soccer}, RL has been successfully applied in learning the keep-away subtask in the RoboCup 2D \cite{AB05} and 3D \cite{Andreas2011}
Soccer Simulation Leagues. Also, in other RoboCup leagues, such as the Middle Size League, RL
has been applied successfully to acquire competitive behaviors \cite{Gabel06bridgingthe}. One
of the noticeable impact on RL is reported by the Brainstormers team, the RoboCup 2D
Simulation League team, on learning different subtasks \cite{DBLP:conf/cig/RiedmillerG07}. A comprehensive analysis of a general batch RL
framework for learning challenging and complex behaviors in robot soccer is reported in
\cite{DBLP:journals/arobots/RiedmillerGHL09}. Despite convergence guarantees, Q($\lambda$)
\cite{sutton98a} with linear function approximation has been used in role assignment in robot soccer
\cite{Kose_2004} and faster learning is observed with the introduction of heuristically accelerated
methods \cite{DBLP:conf/epia/GurzoniTB11}. The dynamic role allocation framework based on dynamic
programming is described in \cite{AAMAS12-MacAlpine} for real-time soccer environments. The role
assignment with this method is tightly coupled with the agent's low-level abilities and does not take
the opponents into consideration. On the other hand, the proposed framework uses the knowledge of the opponent
positions as well as other dynamics for the role assignment function.

Sutton et al. \cite{DBLP:conf/atal/SuttonMDDPWP11} have introduced a real-time learning
architecture, Horde, for expressing knowledge using General Value Functions (GVFs). Our
research is built on Horde to ask a set of questions such that the agents assign optimal or near-optimal
roles within formations. In addition, following researches describe methods and
components to build strategic agents: \cite{ScalingUp2012} describes a methodology to build
a cognizant robot that possesses vast amount of situated, reversible and expressive knowledge.
\cite{DBLP:journals/corr/abs-1112-1133} presents a methodology to ``next'' in real time predicting
thousands of features of the world state, and \cite{conf/smc/ModayilWPS12} presents methods predict about temporally extended consequences of a robot's behaviors in general forms of knowledge. The GVFs are successfully used (e.g.,
\cite{6290309,DBLP:conf/icdl-epirob/WhiteMS12}) for switching and prediction tasks in assistive biomedical robots.

%%%%%%%%%%%%%%%%%%%%%%%%%%%%%%%%%%%%%%%%%%%%%%%%%%%%%%%%%%%%%%%%%%%%%%%%%%%%%%%%%%%%%%%%%%%%%%%%%%

\section{Learnable knowledge representation for Robotic Soccer}
\label{sec:LearnableknowledgeRepresentationForRoboticSoccer}

Recently, within the context of the RL framework \cite{sutton98a}, a
knowledge representation language has been introduced, that is expressive and learnable from sensorimotor
data. This representation is directly usable for robotic soccer as agent-environment interactions
are conducted through perceptors and actuators. In this approach, knowledge is represented as a large
number of \textit{approximate value functions} each with its \begin{inparaenum} \item \textit{own policy};
\item \textit{pseudo-reward function}; \item \textit{pseudo-termination function}; and
\item \textit{pseudo-terminal-reward function}
\cite{DBLP:conf/atal/SuttonMDDPWP11}. \end{inparaenum}
In continuous state spaces, approximate value functions are learned using function approximation and
using more efficient off-policy learning algorithms. First, we briefly introduce  some of the
important concepts related to the GVFs. The complete information about the GVFs are available in
\cite{DBLP:conf/atal/SuttonMDDPWP11,Maei_Sutton_2010,DBLP:conf/icml/MaeiSBS10,MaeiHRPhdThesis2011}.
Second, we show its direct application to simulated robotic soccer.

\subsection{Interpretation}
The interpretation of the approximate value function as a knowledge representation language grounded
on information from perceptors and actuators is defined as:
\begin{definition} \label{def:interpretation}
The knowledge expressed as an \textit{approximate value function} is \textit{true or accurate}, if its numerical
values matches those of the mathematically defined \textit{value function} it is approximating.
\end{definition}

Therefore, according to the Definition (\ref{def:interpretation}), a value function asks a \textit{question}, and an approximate value function is the
\textit{answer} to that question. Based
on prior interpretation, the standard RL framework extends to represent learnable knowledge as
follows. In the standard RL framework \cite{sutton98a}, let the agent and the world interact in discrete
time steps $t=1,2,3,\ldots$. The agent senses the state at each time step $S_t \in \mathcal{S}$, and
selects an action $A_t \in \mathcal{A}$. One time step later the agent receives
a scalar reward $R_{t+1} \in \mathbb{R}$, and senses the state $S_{t+1} \in \mathcal{S}$. The rewards are
generated according to the \textit{reward function} $r:S_{t+1}\rightarrow \mathbb{R}$. The
objective of the standard RL framework is to learn the stochastic action-selection \textit{policy}
$\pi: \mathcal{S} \times \mathcal{A} \rightarrow [0,1]$, that gives the probability of selecting each
action in each state, $\pi(s, a) = \pi(s|a) = \mathcal{P}(A_t = a|S_t = s)$, such that the agent maximizes
rewards summed over the time steps. The standard RL framework extends to include a
\textit{terminal-reward-function}, $z:\mathcal{S} \rightarrow \mathbb{R}$, where $z(s)$ is the terminal
reward received when the termination occurs in state $s$. In the RL framework, $\gamma \in [0,1)$ is used to
discount delayed rewards. Another interpretation of the discounting factor is a constant probability of
$1-\gamma$ termination of arrival to a state with zero terminal-reward. This factor is generalized to
a \textit{termination function} $\gamma:\mathcal{S} \rightarrow [0,1]$, where $1- \gamma(s)$ is the
probability of termination at state $s$, and a terminal reward $z(s)$ is generated.

\subsection{Off-Policy Action-Value Methods for GVFs}
\label{subsec:offpolicyGVFs}
The first method to learn about GVFs, from off-policy experiences, is to use action-value functions. Let $G_t$ be the complete return from state $S_t$ at time $t$, then the sum of the rewards (transient plus
terminal) until termination at time $T$ is:
\[G_t = \sum_{k=t+1}^T r(S_{k}) + z(S_T).\]
The action-value function is:
\[Q^\pi(s,a) = \mathbb{E}(G_t|S_t = s, A_t = a, A_{t+1:T-1}\sim \pi, T \sim \gamma),\]
where, $Q^\pi:\mathcal{S}\times
\mathcal{A}\rightarrow \mathbb{R}$. This is the expected return for a trajectory started from state $s$,
and action $a$, and selecting actions according to the policy $\pi$, until termination occurs
with $\gamma$. We approximate the action-value function with $\hat{Q}:\mathcal{S}\times
\mathcal{A}\rightarrow \mathbb{R}$. Therefore, the action-value function is a precise grounded
question, while the approximate action-value function offers the numerical answer. The complete algorithm for Greedy-GQ($\lambda$) with linear function approximation  for GVFs learning is as shown in Algorithm (\ref{alg:gredyGqLambda}).

\begin{algorithm}
\begin{algorithmic}[1]
    \State \textbf{Initialize} $w_0$ to $0$, and $\theta_0$ arbitrary.
    \State \textbf{Choose} proper (small) positive values for $\alpha_\theta$, $\alpha_w$, and set
values for $\gamma(.) \in (0,1]$, $\lambda(.) \in [0, 1]$.
    \Repeat
         \State \textbf{Initialize} $e=0$.
         \State \textbf{Take} $A_t$ from $S_t$ according to $\pi_b$, and arrive at $S_{t+1}$.
         \State \textbf{Observe} sample, ($S_t, A_t,r(S_{t+1}),z(S_{t+1}), S_{t+1},$) at time step $t$ (with their
corresponding state-action feature vectors), where $\hat{\phi}_{t+1} = \phi(S_{t+1}, A_{t+1}^*),
A_{t+1}^* = \argmax_b {\bf \theta}_t^\mathrm{T} \phi(S_{t+1}, b)$.
         \For{each observed sample}	
           \State $\delta_t \leftarrow r(S_{t+1}) + (1-\gamma(S_{t+1}))z(S_{t+1}) +
\gamma(S_{t+1})
\theta_t^\mathrm{T}
\hat{\phi}_{t+1} - \theta_t^\mathrm{T} \phi_{t}$.
           \State \textbf{If} {$A_t = A_t^*$}, \textbf{then} $\rho_t \leftarrow
\frac{1}{\pi_b(A_t^*|S_t)}$; \textbf{otherwise} $\rho_t \rightarrow 0$.
            \State $e_t \leftarrow I_t \phi_t + \gamma(S_t)\lambda(S_t)\rho_t e_{t-1}$.
            \State $\theta_{t+1} \leftarrow \theta_t + \alpha_\theta[\delta_t e_t - \gamma(S_{t+1})(1
-
\lambda(S_{t+1}))(w_t^\mathrm{T} e_t) \hat{\phi}_{t+1}]$.
            \State $w_{t+1} \leftarrow w_t + \alpha_w [\delta_t e_t - (w_t^\mathrm{T} \phi_t)
\phi_t)]$.
         \EndFor
    \Until{ each episode.}
\end{algorithmic}
\caption{Greedy-GQ($\lambda$) with linear function approximation  for GVFs learning \cite{MaeiHRPhdThesis2011}.}
\label{alg:gredyGqLambda}
\end{algorithm}

The GVFs are defined over four functions: $\pi, \gamma, r,\mbox{and }z$. The functions $r\mbox{ and
}z$ act as pseudo-reward and pseudo-terminal-reward functions respectively. Function $\gamma$ is also in
pseudo form as well. However, $\gamma$ function is more substantive than reward functions as the termination interrupts
the normal flow of state transitions. In pseudo termination, the standard termination is omitted. In
robotic soccer, the base problem can be defined as the time until a goal is scored by either the home or the opponent team. We can consider a pseudo-termination has occurred when the striker is changed.
The GVF with respect to a state-action function is defined as: \[Q^{\pi,\gamma, r,z}(s,a)= \mathbb{E}(G_t|S_t =s, A_t =a,
A_{t+1:T-1}\sim \pi, T
\sim \gamma).\]
The four functions, $\pi, \gamma, r,\mbox{and }z$, are the \textit{question functions}
to GVFs, which in return defines the general value function's semantics. The RL agent learns an approximate
action-value function, $\hat{Q}$, using the four auxiliary functions
$\pi,\gamma, r$ and $z$. We assume that the state space is continuous and the action space is
discrete. We approximate the action-value function using a linear function approximator. We use a
feature extractor $\mathcal{\phi}: S_t \times A_t \rightarrow \{0,1\}^N, N \in \mathbb{N}$, built on tile coding
\cite{sutton98a} to generate feature vectors from state variables and actions. This is a sparse
vector with a constant number of ``1'' features, hence, a constant norm. In addition, tile coding has
the key advantage of real-time learning and to implement computationally efficient algorithms to
learn approximate value functions. In linear function approximation, there exists a weight vector,
$\theta \in \mathbb{R}^N, N \in \mathbb{N}$, to be learned. Therefore, the approximate GVFs are defined as:
\[\hat{Q}(s,a,\theta)=\theta^\mathrm{T}\phi(s,a),\] such that, $\hat{Q}:\mathcal{S} \times \mathcal{A}
\times \mathbb{R}^N \rightarrow \mathbb{R}$. Weights are learned using the gradient-descent temporal-difference
Algorithm (\ref{alg:gredyGqLambda}) \cite{MaeiHRPhdThesis2011}. The Algorithm learns stably and efficiently using linear
function approximation from \textit{off-policy} experiences. Off-policy experiences are generated
from a \textit{behavior policy}, $\pi_b$, that is different from the policy being learned about named
as \textit{target policy}, $\pi$. Therefore, one could learn multiple target policies from the same
behavior policy.

\subsection{Off-Policy Policy Gradient Methods for GVFs}

The second method to learn about GVFs is using the off-policy policy gradient methods with actor-critic architectures that use a state-value function suitable for learning GVFs. It is defined as:
\[
 V^{\pi,\gamma, r,z}(s) = \mathbb{E}(G_t|S_t =s,A_{t:T-1}\sim \pi, T
\sim \gamma),
\]
where, $V^{\pi,\gamma, r,z}(s)$ is the true state-value function, and the approximate GVF is defined as:
\[\hat{V}(s,v)=v^\mathrm{T}\phi(s),\]

where, the functions $\pi,\gamma, r, \mbox{and }z$ are defined as in the subsection (\ref{subsec:offpolicyGVFs}). Since our the target policy $\pi$ is discrete stochastic, we use a Gibbs distribution of the form:
\[
 \pi(a | s) = \frac{e^{u^\mathrm{T} \phi(s, a)}}{\sum_{b}e^{u^\mathrm{T} \phi(s, b)}},
\]

 where, $\phi(s,a)$ are state-action features for state $s$, and action $a$, which are in general unrelated to state features $\phi(s)$, that are used in state-value function approximation. $u \in \mathbb{R}^{N_u}, N_u \in \mathbb{N}$, is a weight vector, which is modified by the actor to learn about the stochastic target policy. The log-gradient of the policy at state $s$, and action $a$, is:
\[
 \frac{\nabla_u \pi(a | s)}{\pi(a | s)} = \phi(s,a) - \sum_b \pi(b|s) \phi(s,b).
\]
The complete algorithm for Off-PAC with linear function approximation  for GVFs learning is as shown in Algorithm (\ref{alg:offPACAlgorithm}).

\begin{algorithm}
\begin{algorithmic}[1]
    \State \textbf{Initialize} $w_0$ to $0$, and $v_0$ and $u_0$ arbitrary.
    \State \textbf{Choose} proper (small) positive values for $\alpha_v$, $\alpha_w$, $\alpha_u$, and set
values for $\gamma(.) \in (0,1]$, $\lambda(.) \in [0, 1]$.
    \Repeat
         \State \textbf{Initialize} $e^v=0, \mbox{and } e^u = 0$.
         \State \textbf{Take} $A_t$ from $S_t$ according to $\pi_b$, and arrive at $S_{t+1}$.
         \State \textbf{Observe} sample, ($S_t, A_t,r(S_{t+1}),z(S_{t+1}), S_{t+1}$) at time step $t$ (with their
corresponding state ($\phi_t, \phi_{t+1}$) feature vectors, where $\phi_t = \phi(S_t)$).
         \For{each observed sample}	
            \State $\delta_t \leftarrow r(S_{t+1}) + (1-\gamma(S_{t+1}))z(S_{t+1}) + \gamma(S_{t+1}) v_t^\mathrm{T} \phi_{t+1} - v_t^\mathrm{T} \phi_{t}$.
            \State $\rho_t \leftarrow \frac{\pi(A_t |S_t)}{\pi_b(A_t|S_t)}$.
            \State Update the critic (GTD($\lambda$) algorithm for GVFs).
            \State \hspace{5mm} $e^v_t \leftarrow \rho_t(\phi_t + \gamma(S_t)\lambda(S_t)e^v_{t-1})$.
            \State \hspace{5mm} $v_{t+1} \leftarrow v_t + \alpha_v[\delta_t e^v_t - \gamma(S_{t+1})(1
-
\lambda(S_{t+1}))({e^v_t}^\mathrm{T} w_t) \phi_{t+1}]$.
            \State \hspace{5mm} $w_{t+1} \leftarrow w_t + \alpha_w [\delta_t e_t - (w_t^\mathrm{T} \phi_t)
\phi_t)]$.
           \State Update the actor.
           \State \hspace{5mm} $e^u_t \leftarrow \rho_t \left[ \frac{\nabla_u \pi (A_t | S_t)}{\pi(A_t | S_t)} + \gamma(S_t) \lambda(S_{t+1}) e^u_{t-1}\right]$.
           \State \hspace{5mm} $u_{t+1} \leftarrow u_t + \alpha_u \delta_t e^u_t$.
         \EndFor
    \Until{ each episode.}

\end{algorithmic}
\caption{Off-PAC with linear function approximation for GVFs learning  \cite{MaeiHRPhdThesis2011,DBLP:journals/corr/abs-1205-4839}.}
\label{alg:offPACAlgorithm}
\end{algorithm}

We are interested in finding optimal policies for the dynamic role assignment, and  henceforth, we use
Algorithms (\ref{alg:gredyGqLambda}), and (\ref{alg:offPACAlgorithm}) for control purposes\footnote{We use an C++ implementation of Algorithm (\ref{alg:gredyGqLambda}) and (\ref{alg:offPACAlgorithm}) in all of our experiments. An implementation is available in \url{https://github.com/samindaa/RLLib}}. We
use linear function approximation for continuous state spaces, and discrete actions are used within
options. Lastly, to summarize, the definitions of the question functions and the answer functions are given as:
\begin{definition}
The question functions are defined by:
\begin{enumerate}
 \item $\pi:S_t \times A_t \rightarrow [0, 1]$ \tabto{35mm} (target policy
is greedy w.r.t. learned value function);
\item $\gamma:S_t \rightarrow [0, 1]$ \tabto{35mm} (termination function);
\item $r:S_{t+1} \rightarrow \mathbb{R}$ \tabto{35mm} (transient reward function); and
\item $z:S_{t+1} \rightarrow \mathbb{R}$ \tabto{35mm} (terminal reward function).
\end{enumerate}
\end{definition}
\begin{definition}
The answer functions are defined by:
\begin{enumerate}
\item $\pi_b: S_t \times A_t \rightarrow [0, 1]$ \tabto{35mm} (behavior policy);
\item $I_t:S_t \times A_t \rightarrow [0, 1]$ \tabto{35mm} (interest function);
\item $\phi:S_t \times A_t \rightarrow \mathbb{R}^N$ \tabto{35mm} (feature-vector function); and
\item $\lambda:S_t \rightarrow [0, 1]$ \tabto{35mm} (eligibility-trace decay-rate function).
\end{enumerate}
\end{definition}

%%%%%%%%%%%%%%%%%%%%%%%%%%%%%%%%%%%%%%%%%%%%%%%%%%%%%%%%%%%%%%%%%%%%%%%%%%%%%%%%%%%%%%%%%%%%%%%%%%
\section{Dynamic Role Assignment}
\label{sec:DynamicRoleAssignment}

A \textit{role} is a specification of an internal or an external behavior of an agent. In our soccer
domain, roles select behaviors of agents based on different reference criteria: the agent close
to the ball becomes the striker. Given a role space, $\mathcal{R}=\{r_1, \ldots, r_n\}$, of size $n$, the
collaboration among $m \leq n$ agents, $\mathcal{A}=\{a_1, \dots, a_m\}$, is obtained through
\textit{formations}. The role space consists of active and reactive roles. For example, the striker is an
active role and the defender could be a reactive role. Given a reactive role, there is a function, $R
\mapsto T$, that maps roles to target positions, $T$, on the field. These target positions are
calculated with respect to a reference pose (e.g., ball position) and other auxiliary criteria
such as crowd management rules. A role assignment function, $R \mapsto A$, provides a
mapping from role space to agent space, while maximizing some reward function. The role assignment
function can be static or dynamic. Static role assignments often provide inferior performance in
robot soccer \cite{AAMAS12-MacAlpine}. Therefore, we learn a dynamic role assignment function
within the RL framework using off-policy control.

\begin{figure}[!h]
\centering
\includegraphics[width=.6\textwidth]{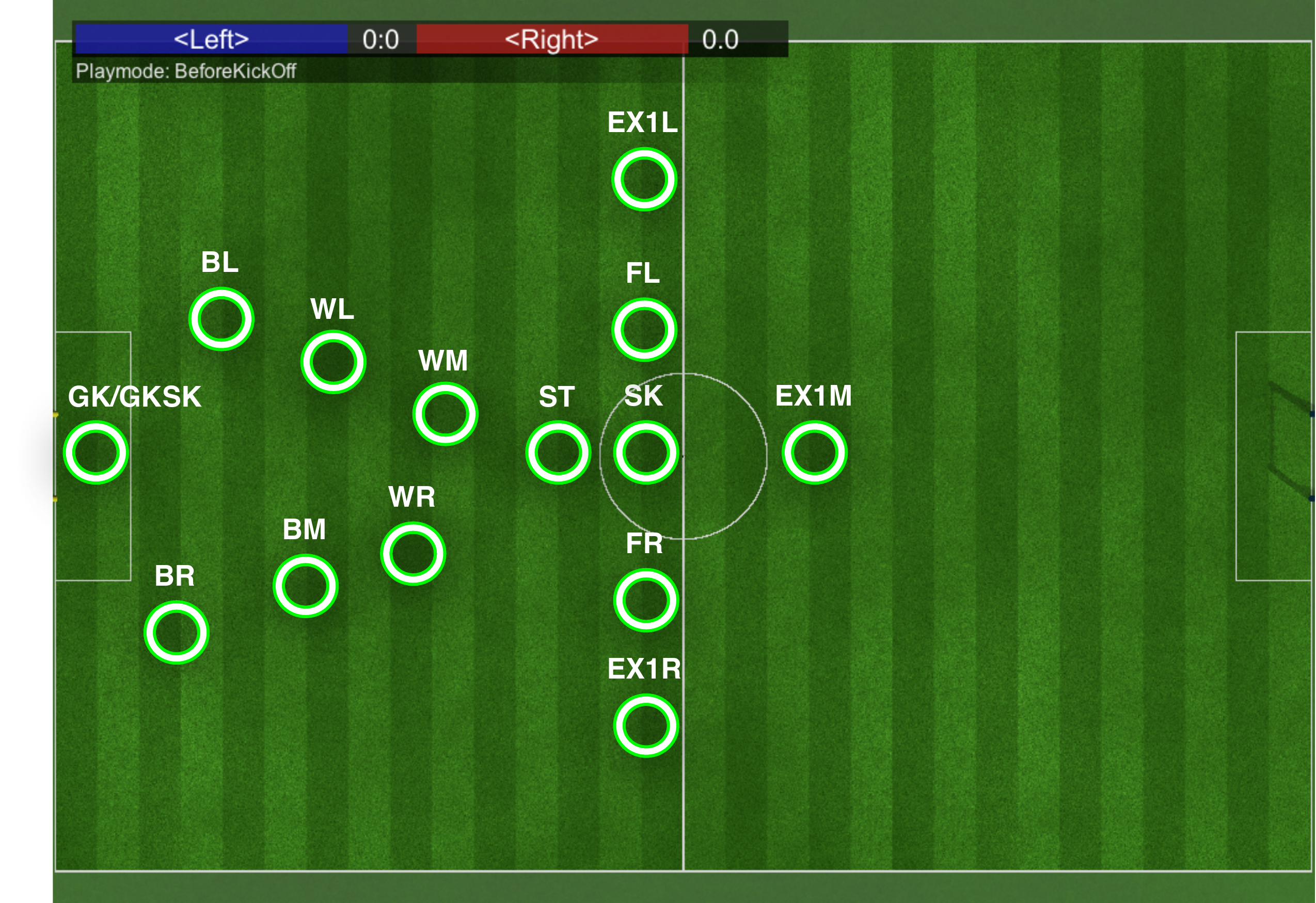}
\caption{Primary formation, \protect\cite{StoeckerV11}}
\label{fig:primaryFormation}
\end{figure}

\subsection{Target Positions with the Primary Formation}

Within our framework, an agent can choose one role among thirteen roles. These roles are part of a
primary formation, and an agent calculates the respective target positions according to its belief of
the absolute ball position and the rules imposed by the 3D soccer simulation
server. We have labeled
the role space in order to describe the behaviors associated with them. Figure
(\ref{fig:primaryFormation}) shows the target positions for the role space before the kickoff state.
The agent closest to the ball takes the striker role ({\sf SK}), which is the only active role.
Let us assume that the agent's belief of the absolute ball position is given by $(x_b,y_b)$. Forward left ({\sf FL})
and forward right ({\sf FR}) target positions are offset by $(x_b,y_b) \pm (0,2)$. The extended
forward left ({\sf EX1L}) and extended forward right (({\sf EX1R})) target positions are offset by
$(x_b,y_b) \pm (0,4)$. The stopper ({\sf ST}) position is given by $(x_b-2.0,y_b)$. The extended
middle ({\sf EX1M}) position is used as a blocking position and it is calculated based on the
closest opponent to the current agent. The other target positions, wing left ({\sf WL}), wing right
({\sf WR}), wing middle ({\sf WM}), back left ({\sf BL}), back right ({\sf BR}), and back middle
({\sf BM}) are calculated with respect to the vector from the middle of the home goal to the ball and offset
by a factor which increases close to the home goal.  When the ball is within the reach of goal keeper,  the
({\sf GK}) role is changed to goal keeper striker ({\sf GKSK}) role. We slightly change the positions when
the ball is near the side lines, home goal, and opponent goal. These adjustments are made in order
to keep the target positions inside the field. We allow target positions to be overlapping. The
dynamic role assignment function may assign the same role during the learning period. In order to
avoid position conflicts an offset is added; the feedback provides negative rewards for such
situations.

\subsection{Roles to RL Action Mapping}

The agent closest to the ball becomes the striker, and only one agent is allowed to become the
striker. The other agents except the goalie are allowed to choose from twelve roles.
We map the available roles to discrete actions of the RL algorithm. In order to use Algorithm
\ref{alg:gredyGqLambda}, an agent must formulate a question function using a
value function, and the answer function provides the solution as an approximate value function. All
the agents formulate the same question: \textit{What is my role in this formation in order to
maximize future rewards?} All agents learn independently according to the question, while
collaboratively aiding each other to maximize their future reward. We make the assumption that the
agents do not communicate their current role. Therefore, at a specific step, multiple agents may
commit to the same role. We discourage this condition by modifying the question as \textit{What is
my role in this formation in order to maximize future rewards, while maintaining a completely
different role from all teammates in all time steps?}

\begin{figure}[!b]
\centering
\includegraphics[width=.8\textwidth]{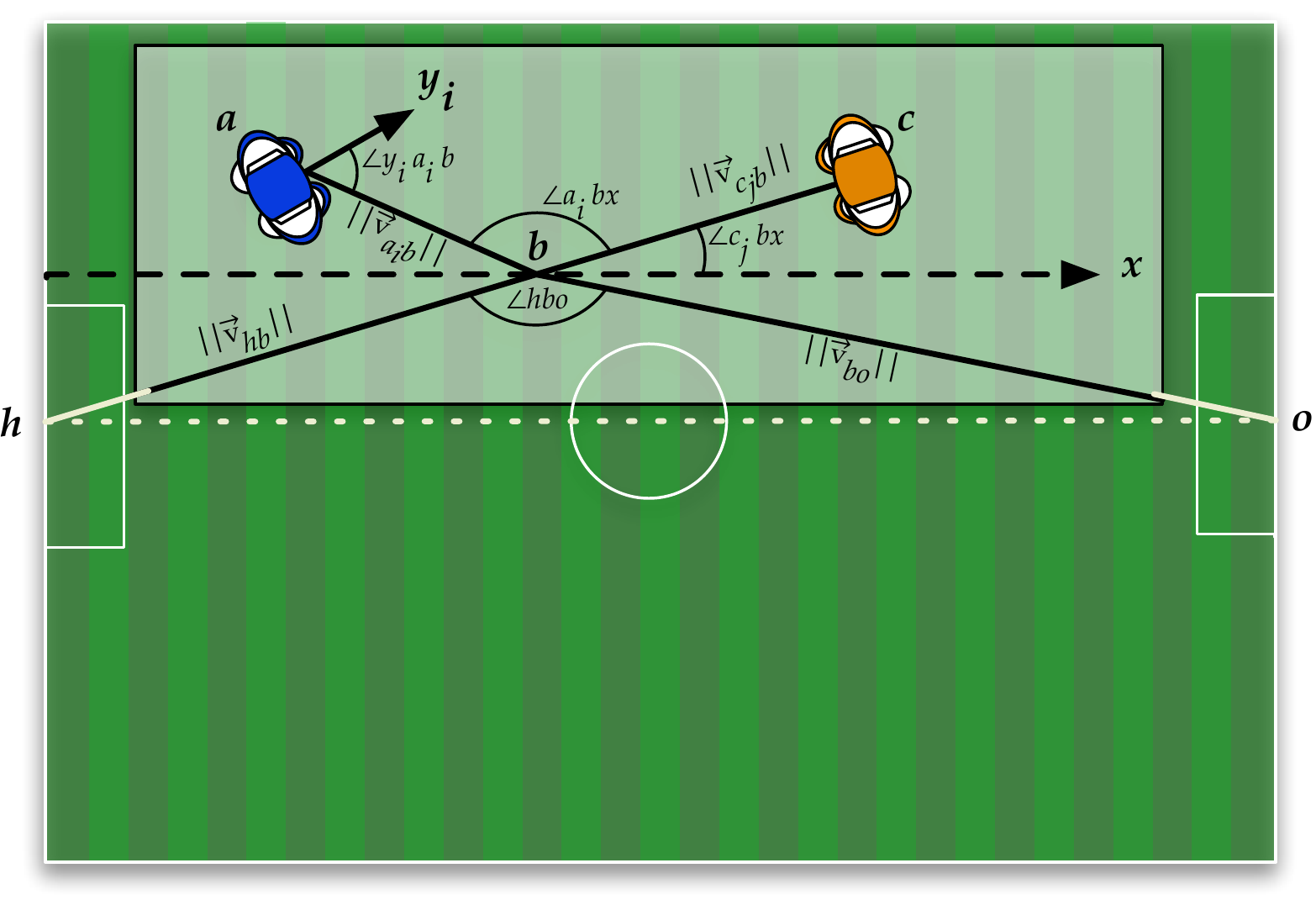}
\caption{State variable representation and the primary function. Some field lines are omitted due to clarity.}
\label{fig:featureRepresentation}
\end{figure}

\subsection{State Variables Representation}

Figure \ref{fig:featureRepresentation} shows the schematic diagram of the state variable
representation. All points and vectors in Figure \ref{fig:featureRepresentation} are defined with
respect to a global coordinate system. $h$ is the middle point of the home goal, while $o$ is the
middle point of the opponent goal. $b$ is the ball position. $\parallel$.$\parallel$ represents the
vector length, while $\angle
pqr$ represents the angle among three points $p,~q,\mbox{ and }r$ pivoted at $q$. $a_i$ represents
the self-localized point of the $i=1,\ldots,11$ teammate agent. $y_i$ is some point in the direction
of the robot orientation of teammate agents. $c_j$, $j=1,\ldots,11$, represents the mid-point of the
tracked opponent agent. $x$ represents a point on a vector parallel to unit vector $e_x$. Using
these labels, we define the state variables as:
\begin{eqnarray*}
  \{\parallel \vec{v}_{hb} \parallel, \parallel \vec{v}_{bo} \parallel, \angle hbo, \{
\parallel
\vec{v}_{a_ib} \parallel, \angle y_i a_i b, \angle a_i b x \}_{i=n_{start}}^{n_{end}},
\{ \parallel
\vec{v}_{c_jb} \parallel, \angle c_j b x, \}_{j=1}^{m_{max}}\}.
\label{eqn:stateVariables}
\end{eqnarray*}

$n_{start}$ is the teammate starting id and $n_{end}$ the ending id. $m_{max}$ is the number of
opponents considered. Angles are normalized to [$-\frac{\pi}{2}, \frac{\pi}{2}$].

\section{Question and Answer Functions}
\label{sec:GVFQandA}

There are twelve actions available in each state. We have left out the striker role from the action set.
The agent nearest to the ball becomes the striker. All agents communicate their belief to other
agents. Based on their belief, all agents calculate a cost function and assign the closest agent as
the striker. We have formulated a cost function based on relative distance to the ball, angle of the
agent, number of teammates and opponents within a region near the ball, and whether the agents are
active. In our formulation, there is a natural termination condition; scoring goals.
With respect to the striker role assignment procedure, we define a pseudo-termination condition. When
an agent becomes a striker, a pseudo-termination occurs, and the striker agent does not participate
in the learning process unless it chooses another role. We define the question and answer
functions as follows:

\subsection{GVF Definitions for State-Action Functions}
\textit{Question functions:}
\begin{enumerate}
\item $\pi=$  greedy w.r.t. $\hat{Q}$,
\item $\gamma(.)=0.8$,
\item $r(.)=$
\begin{inparaenum}
 \item the change of $x$ value of the absolute ball position;
 \item a small negative reward of $0.01$ for each cycle;
 \item a negative reward of $5$ is given to all agents within a radius of 1.5 meters;
\end{inparaenum}
\item $z(.)=$
 \begin{inparaenum}
  \item $+100$ for scoring against opponent;
   \item $-100$ for opponent scoring; and
 \end{inparaenum}
\item $\mbox{time step}= 2$ seconds.
\end{enumerate}

\textit{Answer functions:}
\begin{enumerate}
\item $\pi_b=$  $\epsilon$-greedy w.r.t. target state-action function,
\item $\epsilon=0.05$,
\item $I_t(.)=1$,
\item $\phi(., .)=$
\begin{inparaenum}
 \item we use tile coding to formulate the feature vector.
$n_{start}=2$ and $n_{end}=3,5,7$. $m_{max}=3,5,7$. Therefore, there are $18,28,30$ state variables.
\item state variable is independently tiled with 16 tilings with approximately each with
$\frac{1}{16}$ generalization. Therefore, there are $288+1,448+1,608+1$ active tiles (i.e.,
tiles with feature 1) hashed to a binary vector dimension $10^6+1$. The bias feature is always
active, and
\end{inparaenum}
\item $\lambda(.)=0.8$.
\end{enumerate}

Parameters:\\
\begin{inparaenum}
 \tabto{6mm}\item $\parallel{\bf{\theta}}\parallel=\parallel {\bf w} \parallel = 10^6+1$;
\item $\parallel {\bf e} \parallel=2000$ (efficient trace implementation);
\tabto{6mm} \item $\alpha_\theta=\frac{0.01}{289},\frac{0.01}{449},\frac{0.01}{609}$; and
\item $\alpha_w=0.001\times \alpha_\theta$.
\end{inparaenum}

\subsection{GVF for Gradient Descent Functions}
\textit{Question functions:}
\begin{enumerate}
 \item $\pi=$  Gibbs distribution,
\item $\gamma(.)=0.9$,
\item $r(.)=$
\begin{inparaenum}
 \item the change of $x$ value of the absolute ball position;
 \item a small negative reward of $0.01$ for each cycle;
 \item a negative reward of $5$ is given to all agents within a radius of 1.5 meters;
\end{inparaenum}
\item $z(.)=$
 \begin{inparaenum}
  \item $+100$ for scoring against opponent;
   \item $-100$ for opponent scoring; and
 \end{inparaenum}
\item $\mbox{time step}= 2$ seconds.
\end{enumerate}

\textit{Answer functions:}
\begin{enumerate}
\item $\pi_b=$ the learned Gibbs distribution is used with a small perturbation. In order to provide exploration, with probability $0.01$, Gibbs distribution is perturbed using some $\beta$ value. In our experiments, we use $\beta=0.5$. Therefore, we use a behavior policy: $\frac{e^{u^\mathrm{T} \phi(s, a) + \beta}}{\sum_{b}e^{u^\mathrm{T} \phi(s, b) + \beta}}$
\item $\phi(.)=$
\begin{inparaenum}
 \item the representations for the state-value function, we use tile coding to formulate the feature vector.
$n_{start}=2$ and $n_{end}=3,5,7$. $m_{max}=3,5,7$. Therefore, there are $18,28,30$ state variables.
\item state variable is independently tiled with 16 tilings with approximately each with
$\frac{1}{16}$ generalization. Therefore, there are $288+1,448+1,608+1$ active tiles (i.e.,
tiles with feature 1) hashed to a binary vector dimension $10^6+1$. The bias feature is always
set to active;
\end{inparaenum}
\item $\phi(., .)=$
\begin{inparaenum}
 \item the representations for the Gibbs distribution, we use tile coding to formulate the feature vector.
$n_{start}=2$ and $n_{end}=3,5,7$. $m_{max}=3,5,7$. Therefore, there are $18,28,30$ state variables.
\item state variable is independently tiled with 16 tilings with approximately each with
$\frac{1}{16}$ generalization. Therefore, there are $288+1,448+1,608+1$ active tiles (i.e.,
tiles with feature 1) hashed to a binary vector dimension $10^6+1$. The hashing has also considered the given action. The bias feature is always set to active; and
\end{inparaenum}
\item $\lambda_{\mbox{critic}}(.)=\lambda_{\mbox{actor}}(.)=0.3$.
\end{enumerate}

Parameters:\\
\begin{inparaenum}
\tabto{6mm}\item $\parallel{\bf{u}}\parallel=10^6+1$;
\item $\parallel{\bf{\theta}}\parallel=\parallel {\bf w} \parallel = 10^6+1$;
\tabto{6mm}\item $\parallel {\bf e^v} \parallel=\parallel {\bf e^u} \parallel=2000$ (efficient trace implementation);
\tabto{6mm}\item $\alpha_v=\frac{0.01}{289},\frac{0.01}{449},\frac{0.01}{609}$;
\item $\alpha_w=0.0001\times \alpha_v$; and
\item $\alpha_v=\frac{0.001}{289},\frac{0.001}{449},\frac{0.001}{609}$.
\end{inparaenum}

%%%%%%%%%%%%%%%%%%%%%%%%%%%%%%%%%%%%%%%%%%%%%%%%%%%%%%%%%%%%%%%%%%%%%%%%%%%%%%%%%%%%%%%%%%%%%%%%%

\section{Experiments}
\label{sec:Experiments}

We conducted experiments against the teams {\sf Boldhearts} and {\sf MagmaOffenburg}, both semi-finalists of the RoboCup 3D Soccer Simulation competition in Mexico 2012\footnote{The published binary of the team {\sf UTAustinVilla} showed unexpected behaviors in our tests and is therefore omitted.}. We conducted knowledge learning according to the configuration given in
Section (\ref{sec:GVFQandA}). Subsection (\ref{subsec:expStateActionGVFs}) describes the performance of the Algorithm (\ref{alg:gredyGqLambda}), and Subsection (\ref{subsec:expGradientDescentGVFs}) describes the performance of the Algorithm (\ref{alg:offPACAlgorithm}) for the experiment setup.

\subsection{GVFs with Greedy-GQ($\lambda$)}
\label{subsec:expStateActionGVFs}
\begin{figure}[!b]
\centering
\subfigure[Three vs three agents.] {\includegraphics[width=0.4\textwidth]{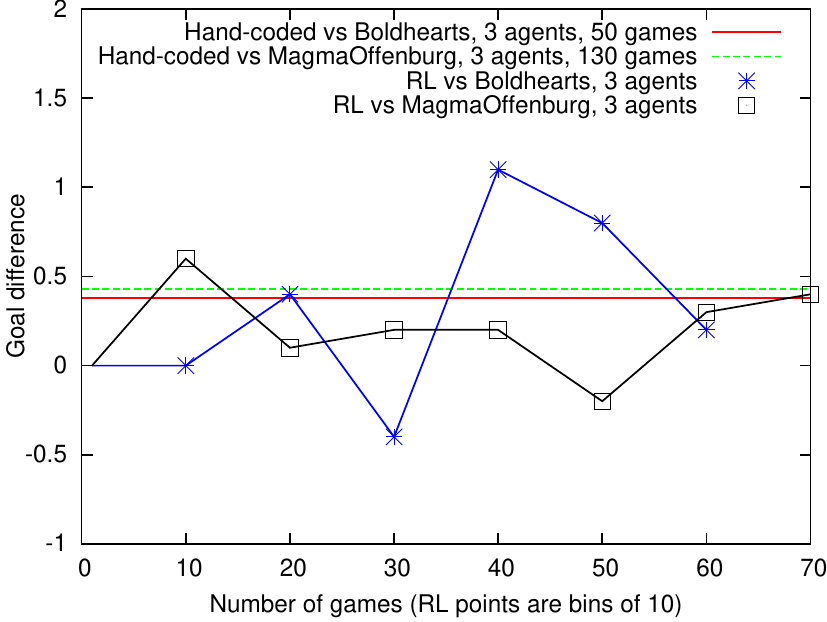} \label{fig:goaldiff3} }
\subfigure[Five vs five agents.] {\includegraphics[width=0.4\textwidth]{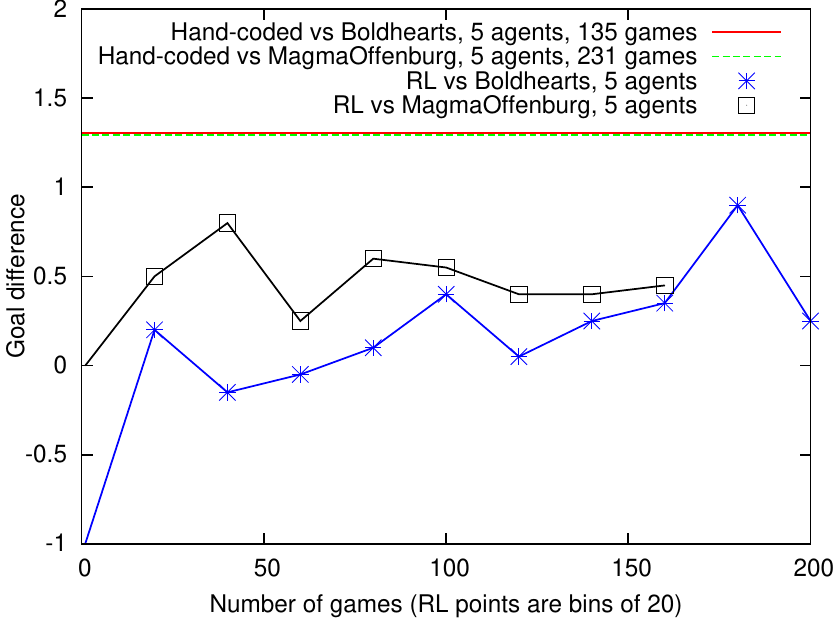} \label{fig:goaldiff5} }
\subfigure[Seven vs seven agents.] {\includegraphics[width=0.4\textwidth]{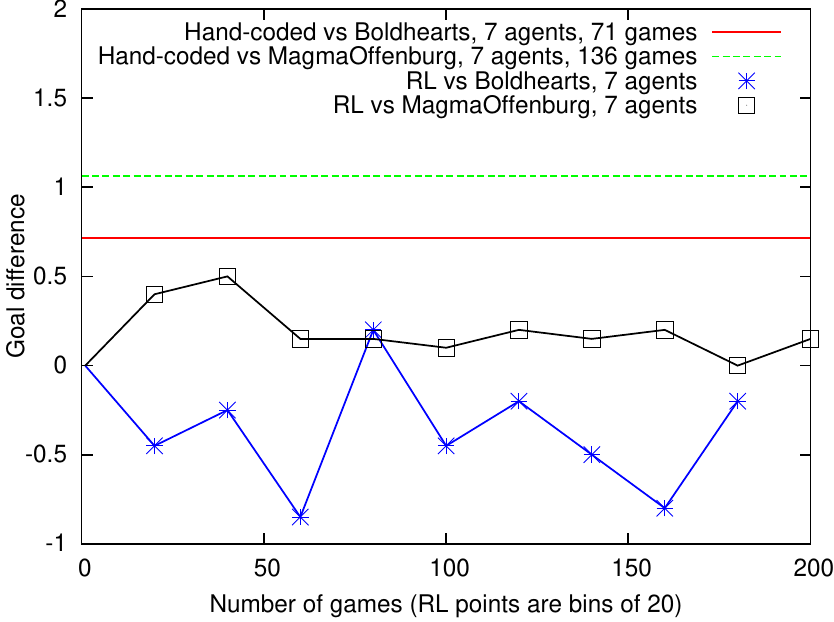} \label{fig:goaldiff7} }
\caption{Goal difference in games with (a) three; (b) five; and (c) seven agents per team using Greedy-GQ($\lambda$) algorithm.}
\label{fig:goaldiffs}
\end{figure}

The first experiments were done using a team size of five with the RL agents against {\sf Boldhearts}. After 140 games our RL agent increased the chance to win from 30\% to 50\%. This number does not increase more in the next games, but after 260 games the number of lost games (initially ~35\%) is reduced to 15\%.
In the further experiments we used the goal difference to compare the performance of the RL agent.
Figure (\ref{fig:goaldiffs}) shows the average goal differences that the hand-tuned role assignment and the RL agents archive in games against {\sf Boldhearts} and {\sf MagmaOffenburg} using different team sizes. With only three agents per team the RL agent only needs 40 games to learn a policy that outperforms the hand-coded role selection (Figure (\ref{fig:goaldiff3})).
Also with five agents per team, the learning agent is able to increase the goal difference against both opponents (Figure (\ref{fig:goaldiff5})). However, it does not reach the performance of the manually tuned role selection. Nevertheless considering the amount of time spent for fine-tuning the hand-coded role selection, these results are promising.
Furthermore, the outcome of the games depends a lot on the underlying skills of the agents, such as walking or dribbling. These skills are noisy, thus the results need to be averaged over many games (std. deviations in Figure (\ref{fig:goaldiffs}) are between 0.5 and 1.3).

The results in Figure (\ref{fig:goaldiff7}) show a bigger gap between RL and the hand-coded agent. However, using seven agents the goal difference is generally decreased, since the defense is easily improved by increasing the number of agents. Also the hand-coded role selection results in a smaller goal difference.
Furthermore, considering seven agents in each team the state space is already increased significantly. Only 200 games seem to be not sufficient to learn a good policy. Sometimes the RL agents reach a positive goal difference, but it stays below the hand-coded role selection.
In Section \ref{sec:ConclusionAndFutureWork}, we discuss some of the reasons for this inferior performances for the team size seven. Even though the RL agent did not perform well considering only the goal difference, it has learned a moderately satisfactory policy. After 180 games the amount of games won is increased slightly from initially 10\% to approximately 20\%.

\subsection{GVFs with Off-PAC}
\label{subsec:expGradientDescentGVFs}

\begin{figure}[!h]
\centering
\subfigure[Three vs three agents.] { \includegraphics[width=0.4\textwidth]{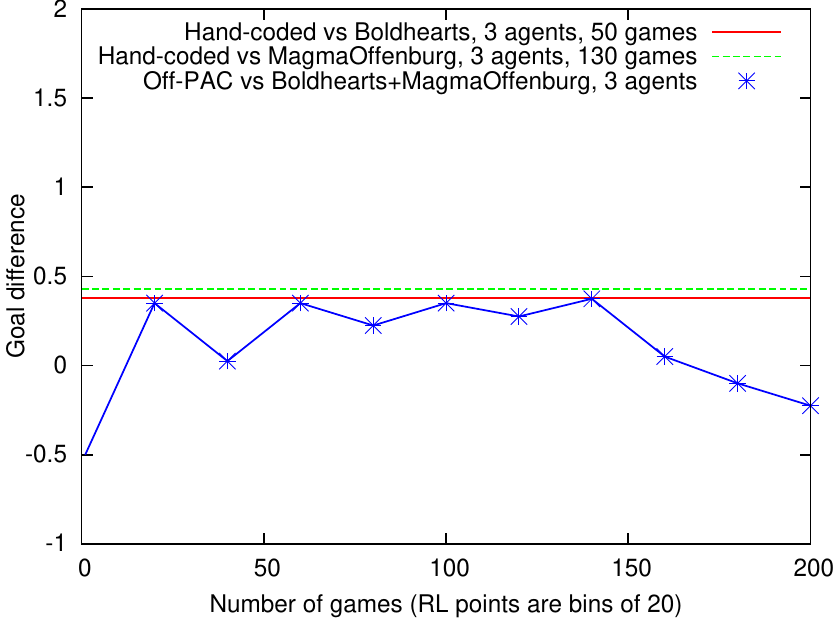} \label{fig:offpac3} }
\subfigure[Five vs five agents.] { \includegraphics[width=0.4\textwidth]{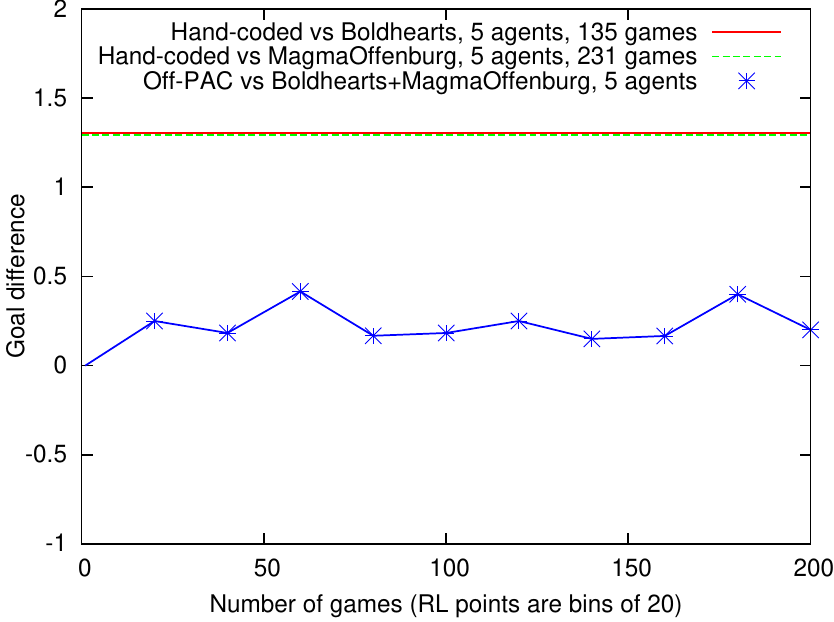} \label{fig:offpac5} }
\subfigure[Seven vs seven agents.] { \includegraphics[width=0.4\textwidth]{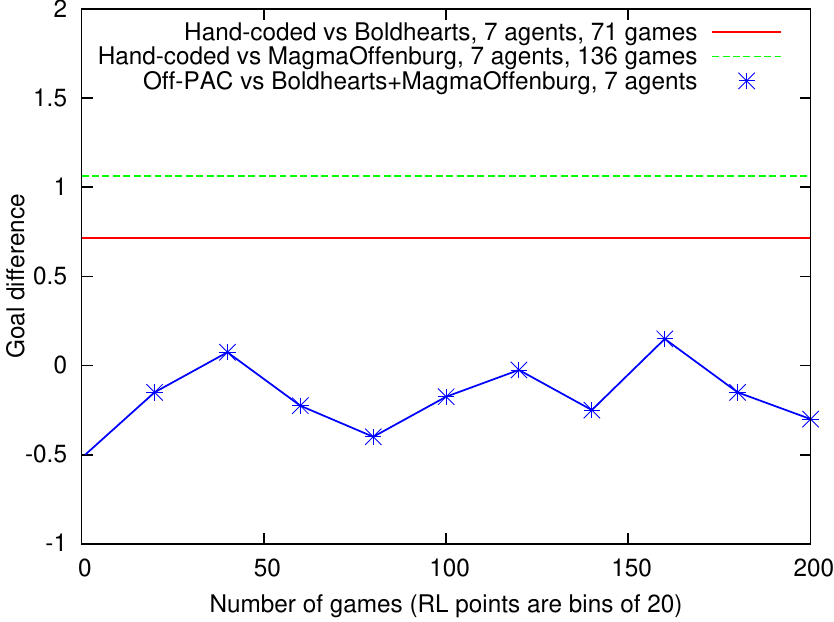} \label{fig:offpac7} }
\caption{Goal difference in games with (a) three; (b) five; and (c) seven agents per team using Off-PAC algorithm.}
\label{fig:goaldiffsOffPAC}
\end{figure}

With Off-PAC, we used a similar environment to that of Subsection (\ref{subsec:expStateActionGVFs}), but with a different learning setup. Instead of learning individual policies for teams separately, we learned a single policy for both teams. We ran the opponent teams in a round robin fashion for 200 games and repeated complete runs for multiple times.   The first experiments were done using a team size of three with RL agents against both teams. Figure (\ref{fig:offpac3}) shows the results of bins of 20 games averaged between two trials. After 20 games, the RL agents have learned a stable policy compared to the hand-tuned policy, but the learned policy bounded above the hand-tuned role assignment function. The second experiments were done using a team size of five with the RL agents against opponent teams. Figure (\ref{fig:offpac5}) shows the results of bins of 20 games averaged among three trials. After 100 games, our RL agent increased the chance of winning to 50\%. This number does not increase more in 
the next games. As Figures (\ref{fig:offpac3}) and (\ref{fig:offpac5}) show, the three and five agents per team are able to increase the goal difference against both opponents. However, it does not reach the performance of the manually tuned role selection. Similar to Subsection (\ref{subsec:expStateActionGVFs}), the amount of time spent for fine-tuning the hand-coded role selection, these results are promising, and the outcome of the experiment heavily depends on the underlying skills of the agents.

The final experiments were done using a team size of seven with the RL agents against opponent teams. Figure (\ref{fig:offpac7}) shows the results of bins of 20 games averaged among two trials.
Similar to Subsection (\ref{subsec:expStateActionGVFs}), with seven agents per team, the results in Figure (\ref{fig:offpac7}) show a bigger gap between RL and the hand-tuned agent. However, using seven agents the goal difference is generally decreased, since the defense is easily improved by increasing the number of agents. Also the hand-tuned role selection results in a smaller goal difference. Figure \ref{fig:offpac7} shows an increase in the trend of winning games. As mentioned earlier, only 200 games seem to be not sufficient to learn a good policy. Even though the RL agents reach a positive goal difference, but it stays below the hand-tuned role selection method. Within the given setting, the RL agents have learned a moderately satisfactory policy. Whether the learned policy is satisfactory for other teams needs to be further investigated.

The RoboCup 3D soccer simulation is inherently a dynamic, and a stochastic environment. There is an infinitesimal chance that a given situation (state) may occur for many games. Therefore, it is paramount important that the learning algorithms extract as much information as possible from the training examples. We use the algorithms in the on-line incremental setting, and once the experience is consumed it is discarded. Since, we learned from off-policy experiences, we can save the tuples, $(S_t, A_t, S_{t+1},r(S_{t+1}),\rho_t, z(S_{t+1}))$, and learn the policy off-line. The Greedy-GQ($\lambda$) learns a deterministic greedy policy. This may not be suitable for complex and dynamic environments such as
the RoboCup 3D soccer simulation environment. The Off-PAC algorithm is designed for stochastic environment. The experiment shows that this algorithm needs careful tuning of learning rates and feature selection, as evident from Figure (\ref{fig:offpac3}) after 160 games.

\section{Conclusions}
\label{sec:ConclusionAndFutureWork}

We have designed and experimented  RL agents that learn to assign roles in order to maximize expected
future rewards. All the agents in the team ask the question ``What is my role
in this formation in order to maximize future rewards, while maintaining a completely different role
from all teammates in all time steps?''. This is a goal-oriented question. We use
Greedy-GQ($\lambda$) and Off-PAC to learn experientially grounded knowledge encoded in GVFs. Dynamic role
assignment function is abstracted from all other low-level components such as walking engine,
obstacle avoidance, object tracking etc. If the role assignment function selects a passive role and
assigns a target location, the lower-layers handle this request. If the lower-layers fail to
comply to this request, for example being reactive, this feedback is not provided to the role
assignment function. If this information needs to be included; it should become a part of the state
representation, and the reward signal should be modified accordingly. The target positions for
passive roles are created w.r.t. the absolute ball location and the rules imposed by the 3D soccer
simulation league. When the ball moves relatively quickly, the target locations
change more quickly.  We have given positive rewards only for the forward ball movements. In order to
reinforce more agents within an area close to the ball, we need to provide appropriate
rewards. These are part of reward shaping \cite{Ng:1999:PIU:645528.657613}. Reward shaping should be
handled carefully as the agents may learn sub-optimal policies not contributing to the overall goal.

The experimental evidences show that agents are learning competitive role assignment functions for
defending and attacking. We have to emphasize that the behavior policy is $\epsilon$-greedy with a
relatively small exploration or slightly perturbed around the target policy. It is not a uniformly distributed policy as used in
\cite{DBLP:conf/atal/SuttonMDDPWP11}. The main reason for this decision is that when an adversary is
present with the intention of maximizing its objectives, practically the learning agent may have to
run for a long period to observe positive samples. Therefore, we have used the
off-policy Greedy-GQ($\lambda$) and Off-PAC algorithms for learning goal-oriented GVFs within on-policy control
setting. Our hypothesis is that with the improvements of the functionalities of lower-layers, the role
assignment function would find better policies for the given question and answer functions. Our
next step is to let the RL agent learn policies against other RoboCup 3D soccer simulation league teams. Beside
the role assignment, we also contributed with testing off-policy learning in high-dimensional state
spaces in a competitive adversarial environment. We have conducted experiments with three, five, and seven
agents per team. The full game consists of eleven agents. The next step is to extend learning
to consider all agents, and to include methods that select informative state variables and features.

\bibliographystyle{splncs03}
\bibliography{references}

%%%%%%%%%%%%%%%%%%%%%%%%%%%%%%%%%%%%%%%%%%%%%%%%%%%%%%%%%%%%%%%%%%%%%%%%%%%%%%%

\end{document}